# Orthogonal multifilters image processing of astronomical images from scanned photographic plates

Vasil Kolev*

***Abstract:*** *In this paper orthogonal multifilters for astronomical image processing are presented. We obtained new orthogonal multifilters based on the orthogonal wavelet of Haar and Daubechies. Recently, multiwavelets have been introduced as a more powerful multiscale analysis tool. It adds several degrees of freedom in multifilter design and makes it possible to have several useful properties such as symmetry, orthogonality, short support, and a higher number of vanishing moments simultaneously. Multifilter decomposition of scanned photographic plates with astronomical images is made.*

## INTRODUCTION

The astronomical images are characterized by special image content and by a very large number of bits/pixel (bpp). These data usually contain the objects (e.g. planets, minor planets, stars, nebulae, galaxies, etc.), which present the subject of interest, investigation and cataloguing. The objects are essentially point sources but their large range of intensities does not allow us to consider all of them to belong to a single region. A night sky data is acquired at low brightness, the acquiring process causes image contamination by various types of noises. In image preview systems the photometric accuracy is not of importance. The faint stars are very close to the plate background.

The wavelet transform is considered as one among the best tools to do this job. It allows us to separate the components of an image according to its size. It is a perfect tool for astronomical images. The effects of three compression algorithms on astrometry and photometry, H-compress, FITSPRESS and the video standard JPEG can be found in [3,4].

The polyspline wavelet is a newer image compression method. In [2] the applications to several astronomical images obtained from scanned photographic plates (SFP) are made. Other image compression methods are multifilters. One of the advantages is that a multiwavelet can possess the orthogonality and symmetry simultaneously, while except for the Haar system, a scalar system cannot have two properties at the same time [4]. Furthermore it is possible to construct non overlapping bases with arbitrary approximation order, which is not possible with one scaling function.

The goal of the present paper is a construction of new orthogonal multiwavelets based on orthogonal wavelets and researching on new multifilters for SFP on the astronomical images.

## ORTHOGONAL MULTISCALING AND MULTIWAVELET FILTER

Multiwavelets generated by a finite set of scalar functions, have advantages in comparison to scalar wavelets. Since multiwavelet decompositions produce two low-pass subbands and two-high pass sub-bands in each dimension, the organization of multiwavelet sub bands differs from the scalar wavelet case. During a single level of decomposition using a scalar wavelet transform, the 2-D image data is replaced with four blocks.

Multiwavelets are base of $L^2(R)$ consisting of more than one base function or generator. One of advantages of multiwavelets is that unlike in the case of a single wavelet, the regularity and approximation order can be improved by increasing the number of generators instead of lengthening the support.

These additional generators then provide more flexibility in approximating a given function.

One way to construct a multiwavelet is through MRA, which consists of a nested sequence $V_j \subset V_{j+1}, j \in Z$, of $L^2(R)$ with the property that the closure of their union is $L^2(R)$ and their intersection is the trivial subspace $\{0\}$. Furthermore, each subspaces $V_j$ is spanned by the dyadic dilates and integer translates of a finite set of scaling functions $\{\phi_i : i = 1,2...,r\}$, sometimes also called the **generators of MRA**. Typically, the scaling vector $\{\Phi = (\phi_i,...,\phi_r)^T : i = 1,2...,r\}$ has compact support or decays rapidly enough at infinity. The support of a scaling vector $\Phi$ is defined as the union of supports of its individual components. For $r = 1$ we obtain the classical wavelet systems. The condition that the spaces $V_j$ be nested implies that the scaling vector $\Phi$ satisfies the following **two-scaling matrix dilation equation** or **matrix refinement equation**

$$\Phi(\mathrm{t}) = \sum_{\mathrm{k} \in \mathrm{Z}} \mathrm{H_k} \Phi(2\mathrm{t - k}) \tag{1}$$

where the filter coefficients matrices $\{\mathrm{H_k}\}_{\mathrm{k} \in \mathrm{Z}}$ are $\mathrm{r} \times \mathrm{r}$ matrices satisfying $\sum_{k \in Z} \|\mathrm{H_k}\|_{l^2(R^{r \times r})} < \infty$ or more explicitly

$$\Phi(\mathrm{t}) = \begin{bmatrix} \phi_1 \\ \phi_2 \\ \vdots \\ \phi_r \end{bmatrix} = \sum_{\mathrm{k} \in \mathrm{Z}} \begin{bmatrix} \mathrm{H}_{11}^{\mathrm{k}} & \mathrm{H}_{12}^{\mathrm{k}} & \cdots & \mathrm{H}_{1\mathrm{r}}^{\mathrm{k}} \\ \mathrm{H}_{21}^{\mathrm{k}} & \mathrm{H}_{22}^{\mathrm{k}} & \cdots & \mathrm{H}_{2\mathrm{r}}^{\mathrm{k}} \\ \vdots & \vdots & & \vdots \\ \mathrm{H}_{\mathrm{r}1}^{\mathrm{k}} & \mathrm{H}_{\mathrm{r}2}^{\mathrm{k}} & \cdots & \mathrm{H}_{\mathrm{rr}}^{\mathrm{k}} \end{bmatrix} \begin{bmatrix} \phi_1(2t-k) \\ \phi_2(2t-k) \\ \vdots \\ \phi_r(2t-k) \end{bmatrix} \tag{2}$$

The previous equation is called the **matrix two-scale relation** or the matrix refinement equation of the **multiscaling function** $\Phi$. Denoting the $\Psi$ complement of $V_j$ in $V_{j+1}$ by $W_j$, it can be shown that there exists a multiwavelet $\Psi$, such that $W_j$ is spanned by the dyadic dilates and integer translates of $\Psi$. Moreover a multiwavelet satisfies a two-scale matrix dilation equation of the form

$$\Psi(\mathrm{t}) = \sum_{\mathrm{k} \in \mathrm{Z}} \mathrm{G_k} \Psi(2\mathrm{t} - \mathrm{k}) \tag{3}$$

where $\mathrm{r} \times \mathrm{r}$ matrix coefficients $\{\mathrm{G(k)}\}_{\mathrm{k} \in \mathrm{Z}}$ are in $l^2(R^{r \times r})$. As an example, for two wavelets ($\mathrm{r} = 2$), we can write the multiscaling function as $\Phi(\mathrm{t}) = \begin{bmatrix} \phi_1 \\ \phi_2 \end{bmatrix}$ and multiwavelet function as $\Psi(\mathrm{t}) = \begin{bmatrix} \psi_1 \\ \psi_2 \end{bmatrix}$. The pair $(\mathrm{H},\mathrm{G})$ is called a **multifilter**, $\mathrm{H}$ is a matrix lowpass filter and $\mathrm{G}$ is a matrix highpass filter, respectively. Both $\Phi$ and $\Psi$ have compact support.

This implies that the sums in (1) and (3) are finite and fully $L^2$- orthogonal in terms of filter coefficients:

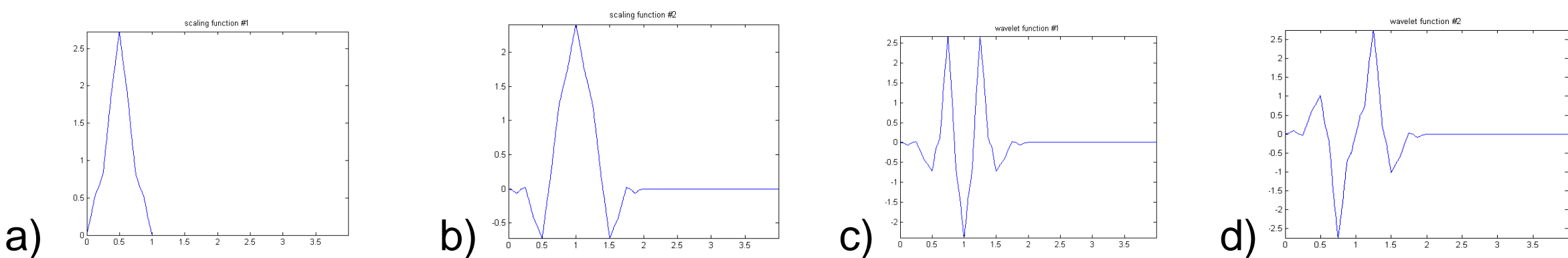

Fig.1 GHM multifilter coefficients, a) multiscaling function $\phi_1(t)$; b) multiscaling function $\phi_2(t)$; c) multiwavelet function $\psi_1(t)$; d) multiwavelet function $\psi_2(t)$

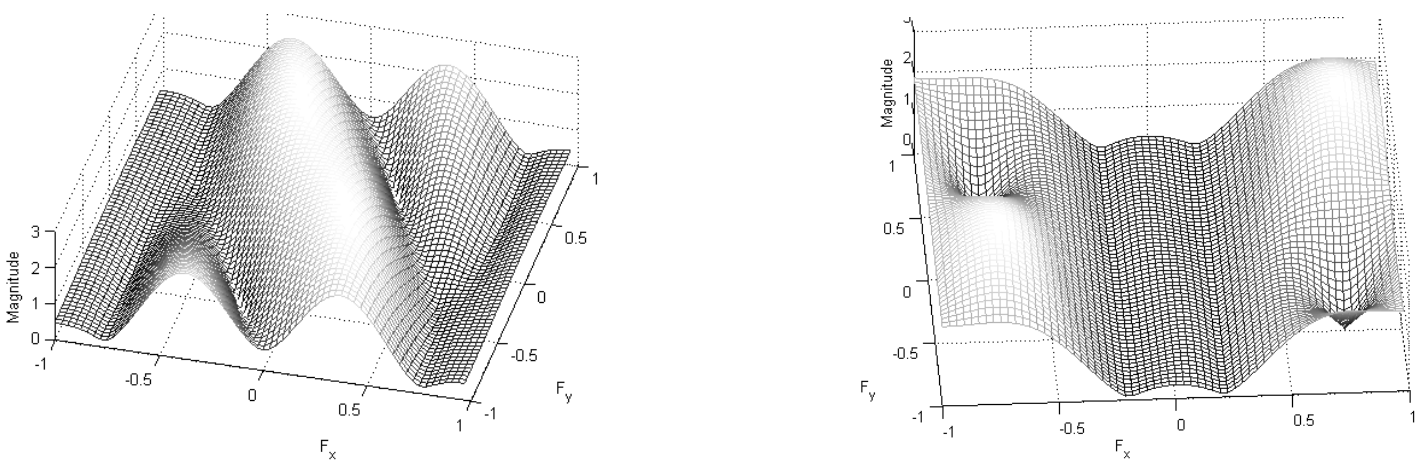

Fig.2 Frequency domain on GHM multifilter; a) Multiscaling filter; b) Multiwavelet filter

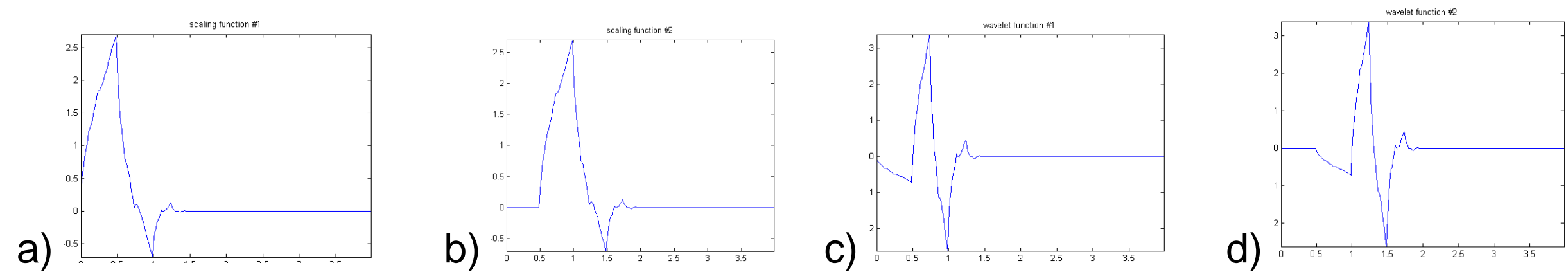

Fig.3 New Daubechies multifilter coefficients, a) multiscaling function $\phi_1(t)$; b) multiscaling function $\phi_2(t)$; c) multiwavelet function $\psi_1(t)$; d) multiwavelet function $\psi_2(t)$

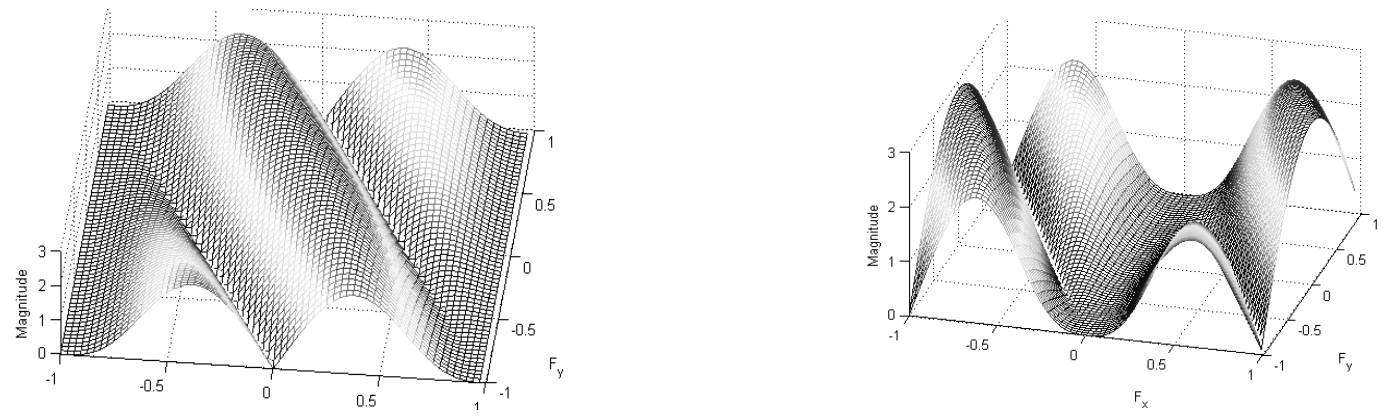

Fig.4 Frequency domain on Daubechies-like multifilter; a) Multiscaling filter $\phi_{1,2}(t)$; b) Multiwavelet filter $\psi_{1,2}(t)$

$$\begin{aligned}
&\sum_{k\in Z} H_k H^T_{k+2l} = \delta_{ol} I_r \\
&\sum_{k\in Z} G_k G^T_{k+2l} = \delta_{ol} I_r \\
&\sum_{k\in Z} H_k G^T_{k+2l} = 0_r \quad , \text{ for all } l \in Z
\end{aligned} \tag{4}$$

Orthogonal multifilter was constructed by, Geronimo, Hardin and Massopust (GHM) [1], Fig.1 and Fig.2, and one level of decomposition, Fig.3.

## DAUBECHIES - LIKE MULTIFILTER

From orthogonal Daubechies wavelet we can construct very nice orthogonal multiwavelets with use double-shift properties. By using (2) for $r = 2$ construct multiscaling filters with $\Phi(t) = \begin{bmatrix} \phi_1(t) \\ \phi_2(t) \end{bmatrix}$ and multiwavelet filters with $\Psi(t) = \begin{bmatrix} \psi_1(t) \\ \psi_2(t) \end{bmatrix}$.

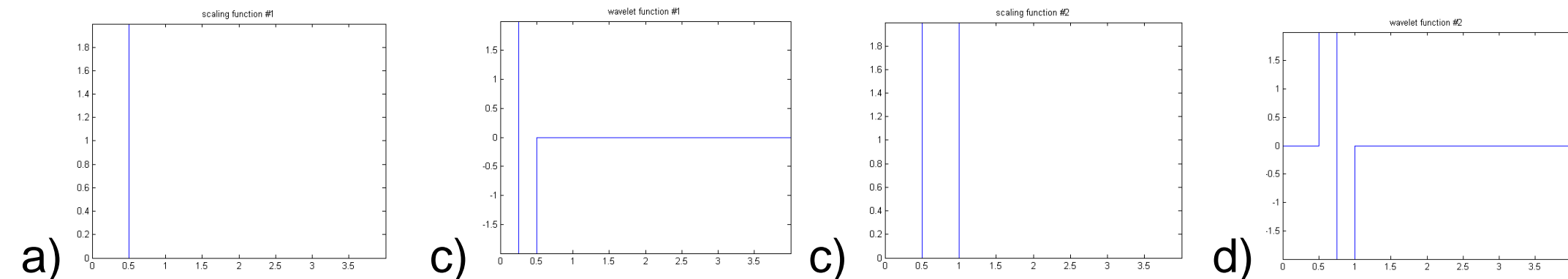

Fig.5 New Haar-like multifilter coefficients, a) Multiscaling function $\phi_1(t)$; b) Multiscaling function $\phi_2(t)$; c) Multiwavelet function $\psi_1(t)$; d) Multiwavelet function $\psi_2(t)$

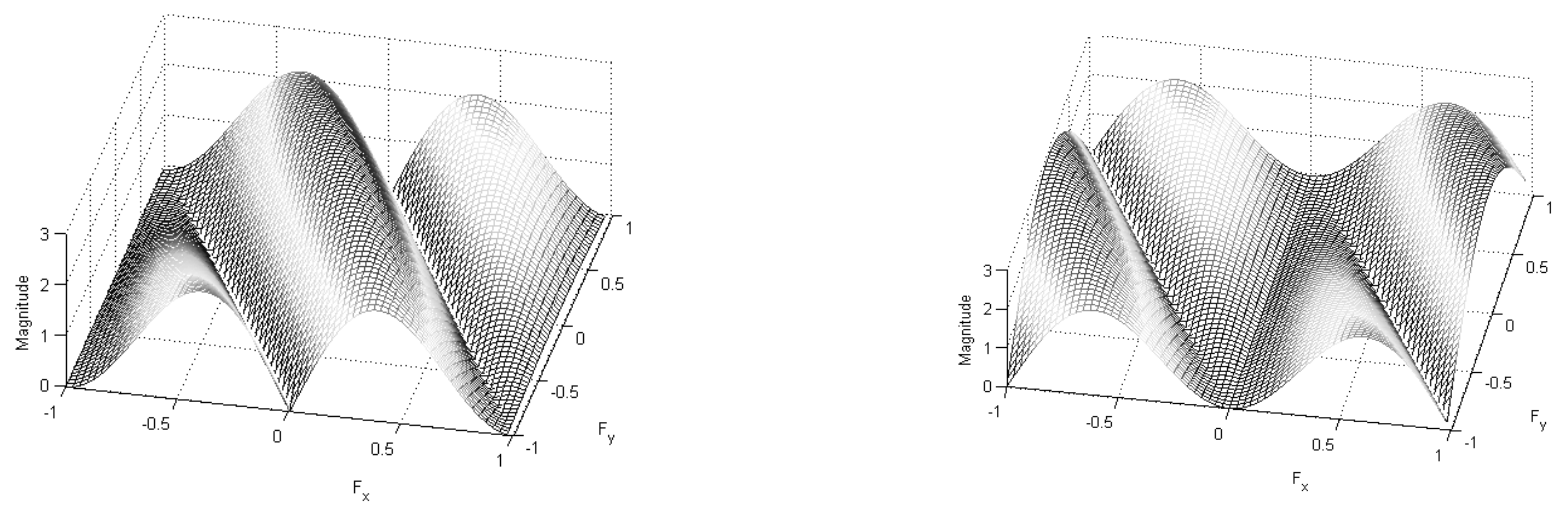

Fig.6 Frequency domain on Haar-like multifilter; a) Multiscaling filter; b) Multiwavelet filter

Therefore, from $\phi_1(t) = \phi(t)$ and $\phi_2(t) = \phi_1(t-2)$ follows that orthogonal multiscaling filter is $\Phi(\mathrm{t}) = \begin{bmatrix} c_0 & c_1 \\ 0 & 0 \end{bmatrix}\Phi(2\mathrm{t}) + \begin{bmatrix} c_2 & c_3 \\ c_0 & c_1 \end{bmatrix}\Phi(2\mathrm{t}-1) + \begin{bmatrix} 0 & 0 \\ c_2 & c_3 \end{bmatrix}\Phi(2\mathrm{t}-2)$ and orthogonal multiwavelet filter $\Psi(\mathrm{t}) = \begin{bmatrix} d_0 & d_1 \\ 0 & 0 \end{bmatrix}\Phi(2\mathrm{t}) + \begin{bmatrix} d_2 & d_3 \\ d_0 & d_1 \end{bmatrix}\Phi(2\mathrm{t}-1) + \begin{bmatrix} 0 & 0 \\ d_2 & d_3 \end{bmatrix}\Phi(2\mathrm{t}-2)$. Hence they satisfy of the orthogonal properties (4). The Daubechies scaling function for four filter coefficients is $\phi(t) = c_0\phi(2t) + c_1\phi(2t-1) + c_2\phi(2t-2) + c_3\phi(2t-3)$ with coefficients $c_0 = (1+\sqrt{3})/4\sqrt{2}$, $c_1 = (3+\sqrt{3})/4\sqrt{2}$, $c_2 = (3-\sqrt{3})/4\sqrt{2}$, and $c_3 = (1-\sqrt{3})/4\sqrt{2}$. The orthogonal multiscaling filter coefficients in time domain are $\mathrm{H}_0 = \begin{bmatrix} c_0 & c_1 \\ 0 & 0 \end{bmatrix}$; $\mathrm{H}_1 = \begin{bmatrix} c_2 & c_3 \\ c_0 & c_1 \end{bmatrix}$; $\mathrm{H}_2 = \begin{bmatrix} 0 & 0 \\ c_2 & c_3 \end{bmatrix}$ and orthogonal multiwavelet coefficients are $\mathrm{G}_0 = \begin{bmatrix} d_0 & d_1 \\ 0 & 0 \end{bmatrix}$; $\mathrm{G}_1 = \begin{bmatrix} d_2 & d_3 \\ d_0 & d_1 \end{bmatrix}$; $\mathrm{G}_2 = \begin{bmatrix} 0 & 0 \\ d_2 & d_3 \end{bmatrix}$. We show the multifilter on Fig.3, and its frequency domain filter coefficients, Fig.4.

**HAAR - LIKE MULTIFILTER**

This multifilter with Haar function was obtained by scaling function $\phi(t) = c_0\phi(2t) + c_1\phi(2t-1)$, with coefficients $c_0 = 1/\sqrt{2}$ and $c_1 = 1/\sqrt{2}$. Therefore we have orthogonal multiscaling filter $\Phi(\mathrm{t}) = \begin{bmatrix} c_0 & c_1 \\ 0 & 0 \end{bmatrix}\Phi(2\mathrm{t}) + \begin{bmatrix} 0 & 0 \\ c_0 & c_1 \end{bmatrix}\Phi(2\mathrm{t}-1)$ and orthogonal multiwavelet filter $\Psi(\mathrm{t}) = \begin{bmatrix} d_0 & d_1 \\ 0 & 0 \end{bmatrix}\Phi(2\mathrm{t}) + \begin{bmatrix} 0 & 0 \\ d_0 & d_1 \end{bmatrix}\Phi(2\mathrm{t}-1)$. Multiscaling filter coefficients are $\mathrm{H}_0 = \begin{bmatrix} c_0 & c_1 \\ 0 & 0 \end{bmatrix}$; $\mathrm{H}_1 = \begin{bmatrix} 0 & 0 \\ c_0 & c_1 \end{bmatrix}$, and the multiwavelet coefficients are $\mathrm{G}_0 = \begin{bmatrix} d_0 & d_1 \\ 0 & 0 \end{bmatrix}$; $\mathrm{G}_1 = \begin{bmatrix} 0 & 0 \\ d_0 & d_1 \end{bmatrix}$. Graphics of multifilter in time domain are presented in Fig.5, and frequency domain – in Fig.6. Here we provide the results with a scan of a plate from the National Astronomical

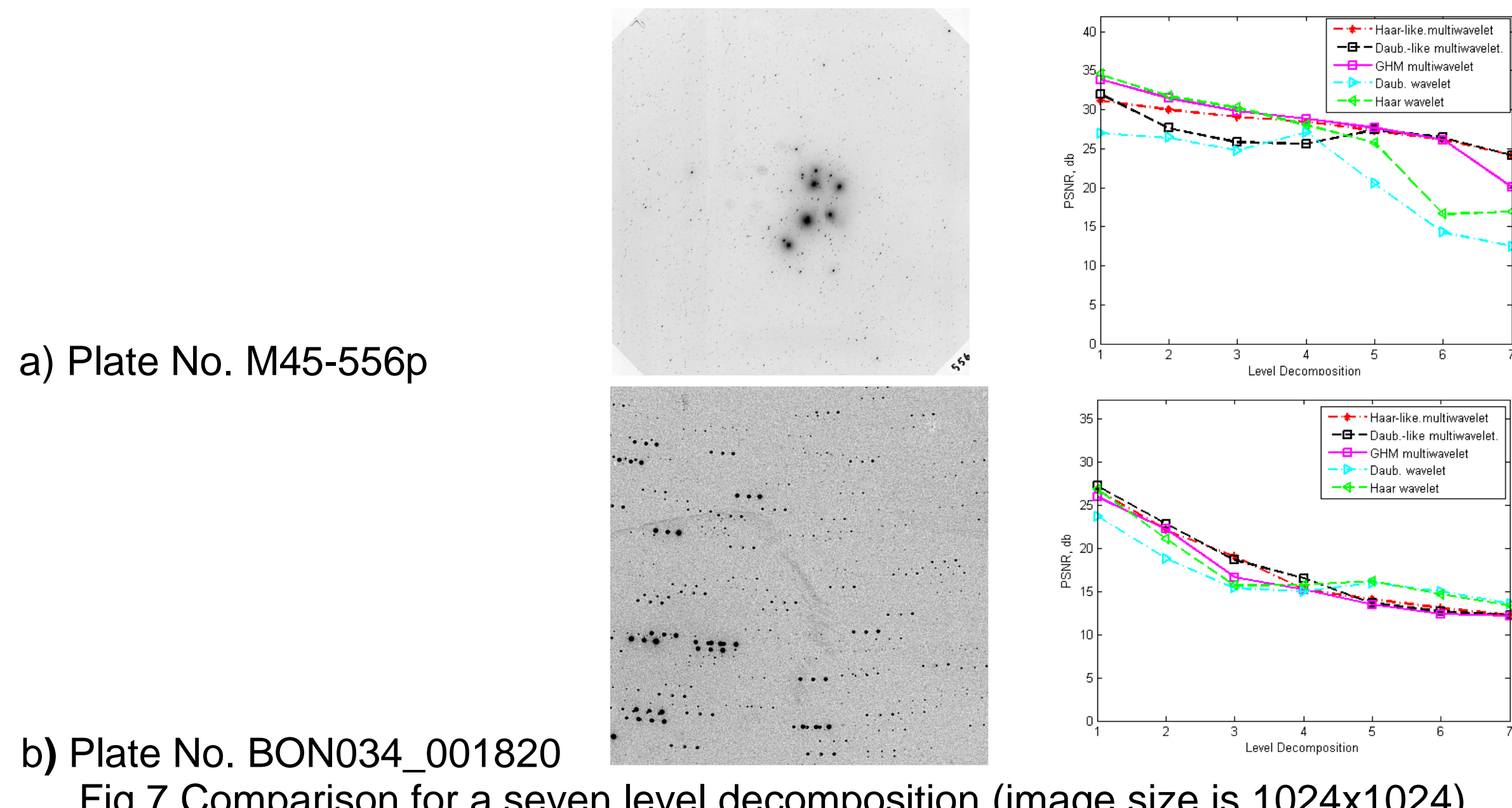

a) Plate No. M45-556p

b) Plate No. BON034_001820

Fig.7 Comparison for a seven level decomposition (image size is 1024x1024)

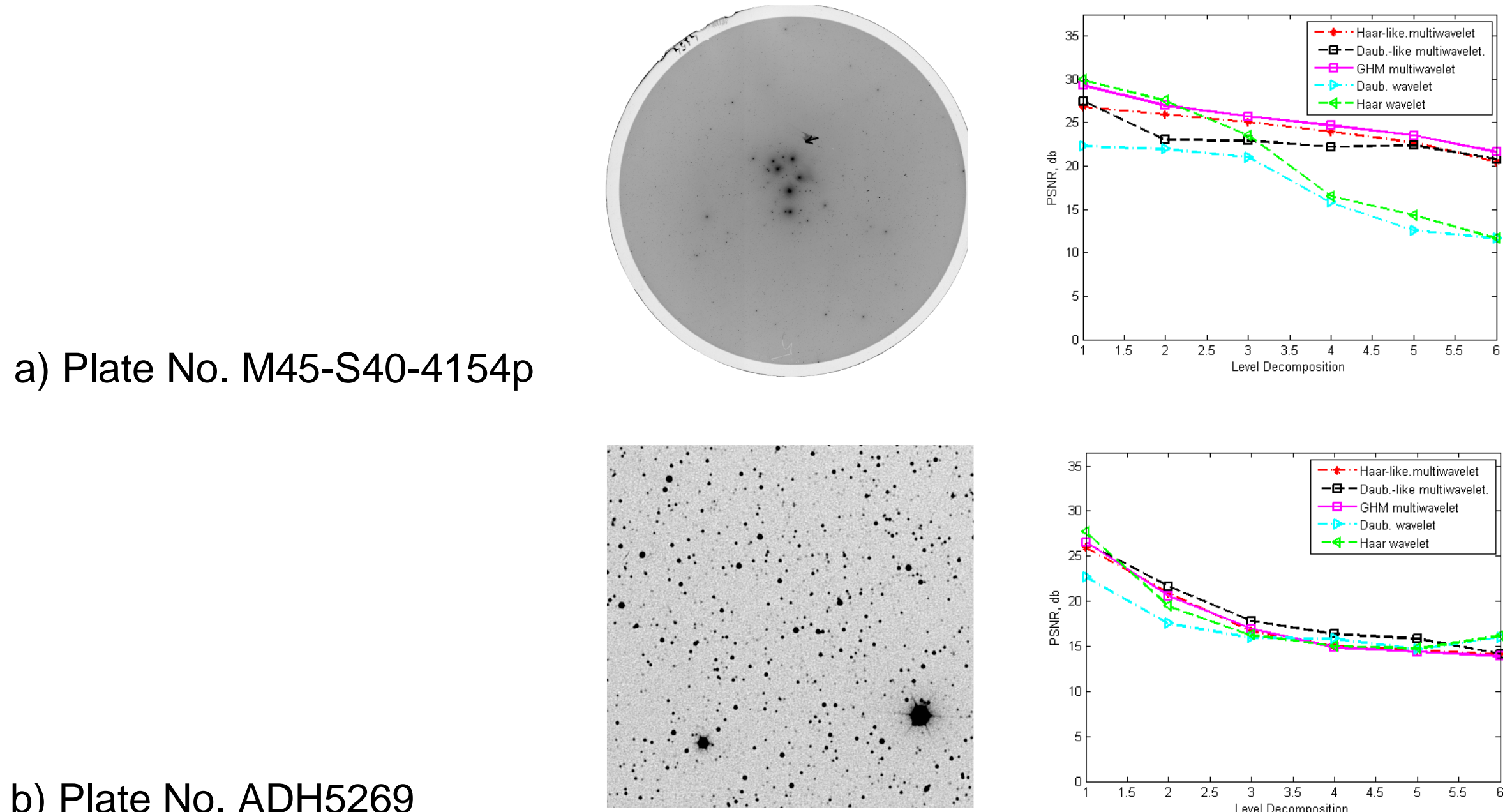

a) Plate No. M45-S40-4154p

b) Plate No. ADH5269

Fig.8 Comparison for a six level decomposition (image size is 512x512)

Observatory Rozhen (Bulgaria). The experiments consider here use SFP on astronomical images with pixel resolution 16 bits. The image quality after applying GHM, new Haar, and new Daubechies multifilters is measured by the peak-signal-to-noise-ratio (PSNR), defined as $\mathrm{PSNR} = 10\log_{10}\left(\frac{255\times 255}{\sqrt{\mathrm{MSE}}}\right), \mathrm{dB}$ and the MSE defined $\mathrm{MSE} = \frac{1}{m\times n}\sum_{i=1}^{m}\sum_{j=1}^{n}(x(i,j)-y(i,j))^2$, where $x(i,j)$, $y(i,j)$ indicate the pixel values in the position $(i.j)$ of original image and that of reconstructed image, respectively, $m, n$ are the width and the height of image. We contained PSNR for every decomposition level on astronomical images, Fig.7, and Fig.8.

## CONCLUSIONS

Multiwavelets have several advantages in comparison with scalar wavelets such as compact and short support, orthogonality, symmetry, and high order approximation. Here the orthogonal multiwavelet filters obtained from orthogonal wavelets were considered with a deferent point of view on the orthogonal multiwavelets. The new multiscaling functions have close frequency domains, see Fig.2a, Fig.4a, and Fig.6a, but multiwavelet functions are different, Fig. 2b, and Fig. 4b. The Haar-like multifilter and Haar scalar transform are lossless transforms, compared with Daubechies-like, Daubechies, and GHM transforms which are lossy transforms.

In Fig.7, and Fig.8, we compare Daubechies-like, Haar-like, and GHM multifilters with Daubechies, and Haar scalar wavelets. We obtain good quality of images and performance with Haar-like multifilters. Compared with the scalar wavelets, Daubechies-like and Haar-like multifilters are with bigger multifilter gains and better choices for the present type of astronomical images. For images characterized by much nonsmooth variation and large regions of uniform intensity, the Haar transforms is extremely effective for lossless compression and it is good choice for up to five level image decompositions, Fig.7a. Disadvantage of Daubechies wavelet is strong dependence on image structures. We can see that at concentration about the image centre on deferent objects, planets, galaxy, pleads, or many stars, is obtained higher quality on images, see Fig.8a and Fig.7a. When we have local mixed object in the image, the present multifilters are a suitable choice to astronomical image processing, because we have large regions of uniform intensity, Fig.7b and Fig.8b. We can use software or hardware implementation of Haar-like multifilters, because they have only two filter coefficients with very simple structure.

In this multifilters image compression the compression performance is dependent on the effectiveness of the decorrelating transform employed. Unfortunately, transform effectiveness is inherently image - dependent, consequently no image transform yield to the best result for all class of images.

**ACKNOWLEDGEMENTS**

I am grateful to Prof. O. Kounchev, from Institute of Mathematics and Informatics, Bulgarian Academy of Sciences, for the advice. I am grateful to Prof. M. Tsvetkov, from Institute of Astronomy, Bulgarian Academy of Sciences, for providing me with FITS images of scanned photographic plates from Sofia Sky Archive Data Center, Bulgarian Academy of Sciences.

This work has been supported by the research project D0-02-275 of the Institutes of Astronomy, and IMI-BAS, Bulgaria

**ABOUT THE AUTHOR**

Vasil Kolev, Institute of Computer and Communications Systems, Bulgarian Academics of Sciences, E-mail: kolev_acad@abv.bg.